\documentclass[11pt]{article}

\usepackage[margin=1in]{geometry}
\usepackage{amsmath,amssymb}
\usepackage{listings}
\usepackage[T1]{fontenc}
\usepackage{xcolor}
\usepackage{tcolorbox}
\usepackage{hyperref}
\usepackage{natbib}

\lstdefinelanguage{Prism}{
  basicstyle=\ttfamily\small,
  columns=fullflexible,
  keepspaces=true,
  backgroundcolor=\color{gray!10},
  frame=single,
  framesep=5pt,
  rulecolor=\color{gray!30}
}

\newtcolorbox{examplebox}[1][]{
  colback=blue!5!white,
  colframe=blue!75!black,
  fonttitle=\bfseries,
  title=#1
}

\title{\textbf{Prism: A Minimal Compositional Metalanguage\\
for Specifying Agent Behaviour}}
\author{
  Franck Binard, PhD\\
  \textit{Deloitte AI}\\
  \and
  Vanja Kljajević, PhD\\
  \textit{University of Oslo}\\
}
\date{}

\begin{document}
\maketitle

\begin{abstract}
This paper introduces \emph{Prism}, a small, compositional metalanguage
for describing the behaviour of software agents that interact with their
environment through external tools (for example, APIs, sensors, or
actuators). The design is deliberately influenced by linguistic notions:
Prism has a tiny, fixed ``core grammar'' and allows users to define
domain-specific ``lexicons'' (contexts) that extend this core with
new categories and predicates.

Instead of hard-wired control structures such as \texttt{if} and
\texttt{else}, Prism uses ordinary expressions and an abstraction
operator \texttt{|} to encode decisions as selections among alternatives.
This makes agent policies look much closer to declarative, grammar-like
specifications than to imperative code, while still being precise enough
to serve as an executable contract for agent behaviour.

We motivate the need for such a language from the perspective of
linguistics and agent design, present the structure of the core context
(\texttt{Core1}), and illustrate how domain-specific contexts can be
used to specify realistic agent behaviours in a fully compositional way.
\end{abstract}

\section{Motivation: Agents as Situated Language Users}

Large language models increasingly act as \emph{agents}: they not only
produce text but also call tools, query databases, and control physical
or virtual devices. From a linguistic point of view, this raises a
familiar problem in a new guise: how do we connect \emph{utterances}
(e.g.~user prompts) to \emph{actions} in a way that is transparent,
compositional, and inspectable?

Two pressures pull in opposite directions:

\begin{itemize}
  \item From engineering, the need for precise, machine-readable
        specifications of what an agent is allowed to do in response
        to different kinds of input (for safety, verification,
        debugging, and explanation).
  \item From linguistics, an interest in systems that respect
        compositional structure, lexicon/grammar separation, and that
        make the link between meaning and action explicit rather than
        opaque.
\end{itemize}

Prism is a proposal aimed at the overlap of these concerns. It is not
a general-purpose programming language; rather, it is a compact
\emph{metalanguage} for writing down agent behaviour in a way that
looks more like a grammar than like a program. Importantly, Prism is 
not a representation language for full natural language semantics; it 
is a control metalanguage that can sit on top of whatever semantic 
representations a model uses internally:

\begin{itemize}
  \item It has a fixed, minimal core (a ``background grammar'').
  \item Domains add their own contexts (mini-grammars) with categories
        such as \texttt{Location}, \texttt{Temperature}, or
        \texttt{Action}.
  \item Decisions are expressed as ordinary expressions that
        \emph{select} between alternatives, rather than as special
        control constructs.
  \item External tools are treated as named capabilities with
        associated schemas, which can be connected directly to
        protocols such as the Model Context Protocol (MCP) \citep{Anthropic2024MCP}, 
        but this complexity is hidden behind a simple surface form.
\end{itemize}

The rest of the paper describes this design, focusing on how it
supports agent behaviour in a way that should be familiar and
interesting to linguists.

\section{Abstraction and the Core Context \texttt{Core1}}

Prism has one main syntactic operator: the vertical bar \texttt{|},
which we use as a generic \emph{abstraction operator}. Informally,
``\texttt{x | body}'' can be read as ``a function from \texttt{x} to 
\texttt{body}'', or if you prefer, ``the meaning of the phrase as a 
function of the argument \texttt{x}''. We also allow multiple
binders, e.g.~\texttt{a, b | body}. This approach draws inspiration
from Church-style abstraction in type theory \citep{Taylor1999}.

The core of Prism is captured as a context called \texttt{Core1}.
Other contexts ``extend'' this core, in the same way that a specific
lexicon might extend a more general one. In what follows, we show
\texttt{Core1} using a concrete notation but describe it informally,
without dwelling on type-theoretic detail.

\subsection{Base Categories and Encoded Structures}

\begin{lstlisting}[language=Prism]
----context
context Core1

Number
String
Unit
Schema
JSON
None
UserPrompt
\end{lstlisting}

These are the basic categories:

\begin{itemize}
  \item \texttt{Number}, \texttt{String}: numerical and textual values.
  \item \texttt{Unit}: measurement units (e.g.~\texttt{celsius}, \texttt{meters}).
  \item \texttt{Schema}, \texttt{JSON}: structures used to describe and
        represent tool inputs and outputs.
  \item \texttt{None}: an empty category used when a function takes no
        meaningful input.
  \item \texttt{UserPrompt}: user utterances or queries.
\end{itemize}

On top of these, \texttt{Core1} defines several structures using
\texttt{|} as an abstraction mechanism.

\paragraph{Booleans.}

\begin{lstlisting}[language=Prism]
Bool := X | X - X - X
\end{lstlisting}

Here, \texttt{Bool} is described as an abstraction over some result
category \texttt{X}. The idea is: a boolean knows how to choose
between two candidates of the same category. Two particular choices
are singled out:

\begin{lstlisting}[language=Prism]
true  := X | a, b | a
false := X | a, b | b
\end{lstlisting}

and from these we can define the usual boolean operations as pure
combinations:

\begin{lstlisting}[language=Prism]
and := b1, b2 | b1[Bool] b2 false
or  := b1, b2 | b1[Bool] true b2
not := b | b[Bool] false true
\end{lstlisting}

The bracket notation \texttt{b[Bool]} means: ``interpret \texttt{b} as
a choice-maker at the category \texttt{Bool}''. More generally, 
\texttt{b[C] a1 a2} can be read as ``use the boolean \texttt{b} to 
choose between \texttt{a1} and \texttt{a2} as values of category 
\texttt{C}''.

\begin{examplebox}[Boolean Operations Example]
\begin{lstlisting}[language=Prism]
-- Combining two predicates
isWarm := gtTemp currentTemp (temp 20 celsius)
isSunny := eq weatherCondition "sunny"
isNice := and isWarm isSunny
\end{lstlisting}
\end{examplebox}

\paragraph{Lists and Predicates.}

\begin{lstlisting}[language=Prism]
List T    := R | (T - R - R) - R - R
Predicate := X | X - X - Bool
\end{lstlisting}

\texttt{List T} encodes the idea of a list over some category
\texttt{T}---it is not used extensively in the examples here, but
forms part of the general core.

\texttt{Predicate} describes binary predicates: for a given category
\texttt{X}, a \texttt{Predicate[X]} is something that, given two
instances of \texttt{X}, yields a \texttt{Bool}. For numbers,
\texttt{Core1} declares standard comparison predicates as \emph{external}
(i.e.~implemented outside the language):

\begin{lstlisting}[language=Prism]
external gt : Predicate[Number]
external lt : Predicate[Number]
external eq : Predicate[Number]
external gte : Predicate[Number]
external lte : Predicate[Number]
\end{lstlisting}

\paragraph{Tools and Pairs.}

\begin{lstlisting}[language=Prism]
Tool     := X | (String - Schema - Schema - X) - X
Pair U V := X | (U - V - X) - X
\end{lstlisting}

A \texttt{Tool} is a triple consisting of a name and input/output
schemas. Intuitively, it is ``a capability the agent can use'', such
as a web API, a database query, or a hardware sensor.

\texttt{Pair U V} encodes a pair of a \texttt{U} and a \texttt{V},
again described in terms of how it interacts with a consumer. The
core introduces a pure constructor:

\begin{lstlisting}[language=Prism]
pair := A, B | a, b | X | x | x a b
\end{lstlisting}

which can be understood as: given \texttt{a} and \texttt{b}, produce a
pair that, when given a consumer \texttt{x}, applies \texttt{x} to
\texttt{a} and \texttt{b}.

\subsection{Core as a Shared Background Grammar}

From a linguistic perspective, \texttt{Core1} supplies a small
``background grammar'':

\begin{itemize}
  \item Categories such as \texttt{Number}, \texttt{String},
        \texttt{Unit}, and \texttt{UserPrompt}.
  \item Basic semantic operators: conjunction, disjunction, negation.
  \item Predicates on numbers: greater-than, less-than, equality, and their
        inclusive variants.
  \item A way to talk about tools (external capabilities).
  \item A way to construct simple product-like structures (pairs).
\end{itemize}

Prism itself does not have keywords like ``if'' or
``otherwise''. Instead, such notions are encoded via these small
combinators.

\section{Contexts as Domain-Specific Mini-Grammars}

A \emph{context} extends \texttt{Core1} with new categories and
external bindings. Syntactically it has the form:

\begin{lstlisting}[language=Prism]
----context
context Name extends Core1

... new categories ...
... definitions (:=) ...
... external tools and actions ...
\end{lstlisting}

From a linguistic perspective, a context is a small domain-specific
grammar: it names entities, introduces roles (such as \texttt{Action}
or \texttt{Location}), and declares which external tools are available
in that domain. However, every context shares the same underlying
combinatorial machinery from \texttt{Core1}.

\paragraph{Notational convention.}
We use capitalised words like \texttt{Location} and \texttt{Action} 
for categories (types), and lowercase words like \texttt{office} or 
\texttt{celsius} for individual values (terms).

\section{Example 1: Thermostat Control}

\subsection{Setting up the Domain}

We begin by defining a context \texttt{ThermostatControl} that extends
\texttt{Core1}.

\begin{lstlisting}[language=Prism]
----context
context ThermostatControl extends Core1

type Location
type Action extends Tool
\end{lstlisting}

\texttt{Location} is a new category for places in the environment
(e.g.~\texttt{office}, \texttt{kitchen}); \texttt{Action} is a role
refining \texttt{Tool}, used specifically for things the agent can do
to the thermostat.

We represent temperatures as number--unit pairs:

\begin{lstlisting}[language=Prism]
Temperature := Pair[Number][Unit]
\end{lstlisting}

Intuitively, \texttt{Temperature} is ``a number together with a unit''. 
Using the core \texttt{pair} constructor, we introduce a convenience
function \texttt{temp} that builds a temperature from a number and a
unit:

\begin{lstlisting}[language=Prism]
temp := n, u | pair[Number][Unit] n u
\end{lstlisting}

For example, \texttt{temp 23 celsius} is the temperature ``23 degrees
Celsius'', or more concisely, the value ``23\,°C''.

We also introduce specific locations and units:

\begin{lstlisting}[language=Prism]
office  : Location
kitchen : Location
bedroom : Location
celsius : Unit
fahrenheit : Unit
\end{lstlisting}

Next, an external sensor:

\begin{lstlisting}[language=Prism]
external tempSensor : Location - Unit - Temperature
\end{lstlisting}

This is the agent's way of reading the temperature. From the
perspective of Prism, \texttt{tempSensor} behaves as something that,
given a location and a unit, yields a temperature value.

Finally, we define external comparisons over temperatures, and the
thermostat actions themselves:

\begin{lstlisting}[language=Prism]
external gtTemp : Predicate[Temperature]
external ltTemp : Predicate[Temperature]

external lowerThermostat : Action
external raiseThermostat : Action
maintainThermostat       : Action
\end{lstlisting}

\texttt{gtTemp} and \texttt{ltTemp} are domain-specific predicates
that say when one temperature is greater or less than another.
\texttt{lowerThermostat}, \texttt{raiseThermostat}, and
\texttt{maintainThermostat} are actions the agent can take.

\subsection{Behavioural Policy as a Prism Expression}

We now express the following policy purely in Prism:

\begin{quote}
If the office temperature is above 23\,°C, lower the
thermostat. Otherwise, if it is below 20.5\,°C, raise the
thermostat. Otherwise, keep the thermostat as it is.
\end{quote}

In Prism:

\begin{lstlisting}[language=Prism]
(gtTemp (tempSensor office celsius) (temp 23 celsius))[Action]
  lowerThermostat
  ((ltTemp (tempSensor office celsius) (temp 20.5 celsius))[Action]
    raiseThermostat
    maintainThermostat)
\end{lstlisting}

The outer boolean chooses between \texttt{lowerThermostat} and the 
entire fallback clause; the inner boolean then chooses between 
\texttt{raiseThermostat} and \texttt{maintainThermostat}.

Read from the top:

\begin{enumerate}
  \item \texttt{tempSensor office celsius} asks the external sensor
        for the temperature in the office, in Celsius.
  \item \texttt{gtTemp (tempSensor office celsius) (temp 23 celsius)}
        is a boolean phrase: ``the office temperature is greater than
        23\,°C''.
  \item Applying this boolean to \texttt{[Action]} and two candidate
        actions:
\begin{lstlisting}[language=Prism]
(gtTemp (...))[Action] lowerThermostat ...
\end{lstlisting}
        yields an \texttt{Action}:
        \begin{itemize}
          \item \texttt{lowerThermostat} if the condition is true;
          \item the ``else'' branch if it is false.
        \end{itemize}
  \item The ``else'' branch is itself another boolean selection:
\begin{lstlisting}[language=Prism]
(ltTemp (tempSensor office celsius) (temp 20.5 celsius))[Action]
  raiseThermostat
  maintainThermostat
\end{lstlisting}
        which chooses between \texttt{raiseThermostat} and
        \texttt{maintainThermostat}, depending on whether the
        temperature is below 20.5\,°C.
\end{enumerate}

Crucially, nowhere do we introduce special syntax for \texttt{if},
\texttt{then}, or \texttt{else}. What looks like an ``if--elseif--else''
structure emerges from the interaction of three components:

\begin{itemize}
  \item domain predicates (\texttt{gtTemp}, \texttt{ltTemp}),
  \item core booleans (\texttt{Bool}, \texttt{true}, \texttt{false},
        and the way they act as selectors),
  \item and a convention of applying booleans with a target category in
        brackets (here \texttt{[Action]}) followed by the two candidate
        results.
\end{itemize}

From a linguistic perspective, the policy above is a kind of
\emph{compositionally specified decision procedure}. Each part is
locally interpretable (e.g.~``\emph{the office temperature is greater
than 23\,°C}'') and combined into a global action choice in
a way that respects a clear scopal structure.

\section{Example 2: Smart Home Security}

\subsection{Domain Context}

\begin{lstlisting}[language=Prism]
----context
context HomeSecurity extends Core1

type Room
type SecurityAction extends Tool
type SecurityLevel

living_room : Room
front_door  : Room
garage      : Room

high   : SecurityLevel
medium : SecurityLevel
low    : SecurityLevel

external motionSensor : Room - Bool
external doorSensor   : Room - Bool
external alertSecurity : SecurityLevel - SecurityAction
external logEvent      : String - SecurityAction
external doNothing     : SecurityAction
\end{lstlisting}

\subsection{Security Policy}

\begin{examplebox}[Security Response Policy]
``If motion is detected in the living room AND the front door is open, 
send a high-level security alert. Otherwise, if only motion is detected,
log the event. Otherwise, do nothing.''

\begin{lstlisting}[language=Prism]
policy := 
  (and 
    (motionSensor living_room) 
    (doorSensor front_door))[SecurityAction]
  (alertSecurity high)
  ((motionSensor living_room)[SecurityAction]
    (logEvent "motion_detected")
    doNothing)
\end{lstlisting}
\end{examplebox}

Note that the nested structure includes a simpler, single-clause decision:
\texttt{(motionSensor living\_room)[SecurityAction] ...} says ``if there 
is motion in the living room, do X, otherwise do Y''. This demonstrates 
how simple conditionals compose into more complex decision trees.

\section{Example 3: E-Commerce Recommendation}

\subsection{Domain Context}

\begin{lstlisting}[language=Prism]
----context
context ECommerce extends Core1

type Customer
type Product
type Recommendation extends Tool

external customerBudget : Customer - Number
external productPrice   : Product - Number
external customerHistory : Customer - List[Product]
external inStock        : Product - Bool

external recommend   : Product - Recommendation
external suggestAlt  : Product - Recommendation
external notifyOutOfStock : Recommendation
\end{lstlisting}

\subsection{Recommendation Policy}

\begin{examplebox}[Product Recommendation Logic]
``If the product is in stock AND the price is within the customer's budget,
recommend it. Otherwise, if the product is out of stock, notify the customer.
Otherwise, suggest an alternative.''

\begin{lstlisting}[language=Prism]
recommendProduct := customer, product |
  (and
    (inStock product)
    (lte (productPrice product) (customerBudget customer))
  )[Recommendation]
  (recommend product)
  ((inStock product)[Recommendation]
    (suggestAlt product)
    notifyOutOfStock)
\end{lstlisting}
\end{examplebox}

\section{Example 4: Medical Alert System}

\subsection{Domain Context}

\begin{lstlisting}[language=Prism]
----context
context MedicalAlert extends Core1

type Patient
type VitalSign := Pair[Number][Unit]
type AlertLevel
type Response extends Tool

bpm    : Unit  -- beats per minute
mmHg   : Unit  -- millimeters of mercury
mgdL   : Unit  -- milligrams per deciliter

critical : AlertLevel
warning  : AlertLevel
normal   : AlertLevel

external heartRate     : Patient - VitalSign
external bloodPressure : Patient - VitalSign
external bloodSugar    : Patient - VitalSign

external emergencyCall  : Patient - AlertLevel - Response
external notifyNurse    : Patient - Response
external logVitals      : Patient - Response

vitalSign := n, u | pair[Number][Unit] n u
\end{lstlisting}

\subsection{Alert Policy}

\begin{examplebox}[Patient Monitoring Policy]
``If heart rate exceeds 120 bpm OR blood pressure systolic exceeds 180 mmHg,
trigger emergency call. Otherwise, if heart rate exceeds 100 bpm, notify nurse.
Otherwise, just log the vitals.''

\begin{lstlisting}[language=Prism]
monitorPatient := patient |
  (or
    (gtTemp (heartRate patient) (vitalSign 120 bpm))
    (gtTemp (bloodPressure patient) (vitalSign 180 mmHg))
  )[Response]
  (emergencyCall patient critical)
  ((gtTemp (heartRate patient) (vitalSign 100 bpm))[Response]
    (notifyNurse patient)
    (logVitals patient))
\end{lstlisting}
\end{examplebox}

\section{Linguistic Perspective}

\subsection{Compositionality and Selection}

Prism is explicitly compositional: each expression is built from
smaller expressions by function application and abstraction. This
aligns well with the Fregean principle of compositionality, that the 
meaning of the whole is a function of the meanings of the parts and 
their mode of combination \citep{Frege1892,Partee1995}.

In the thermostat example, we can identify sub-phrases such as:

\begin{itemize}
  \item \texttt{tempSensor office celsius}: a kind of ``measurement
        phrase''.
  \item \texttt{temp 23 celsius}: a ``degree phrase''.
  \item \texttt{gtTemp (tempSensor office celsius) (temp 23 celsius)}:
        a sentential predicate about the state of the world.
\end{itemize}

\paragraph{From natural language to Prism.}
To illustrate the formalisation role of Prism, consider this mapping:

\begin{center}
\begin{tabular}{p{0.45\textwidth}p{0.45\textwidth}}
\textbf{Natural language} & \textbf{Prism expression} \\
\hline
``If the door is open and motion is detected, send a high-level alert.'' &
\texttt{(and (doorSensor front\_door) (motionSensor living\_room))[SecurityAction] (alertSecurity high) ...}
\end{tabular}
\end{center}

Instead of mapping directly to truth conditions, however, Prism maps
such predicates to \emph{choices over actions}. Booleans, when
instantiated at a category like \texttt{Action}, become selection
functions:

\begin{quote}
Given two alternative actions, choose one.
\end{quote}

This is reminiscent of work in dynamic semantics and decision theory,
where propositions can be seen as tests or update functions, rather
than as static truth values \citep{Groenendijk1991,VanEijck1996}. 
Prism embeds this idea at the level of a very small, concrete notation.
In this view, Prism expressions corresponding to conditionals are 
closer to update rules on an agent's action state than to static 
propositions; they encode how an utterance constrains subsequent 
behaviour.

\subsection{Lexicon vs.\ Grammar}

Contexts in Prism play a role similar to that of lexicons in
generative grammar \citep{Chomsky1995}. The core (\texttt{Core1}) provides the abstract
combinatorial machinery and a few very general categories. A context
such as \texttt{ThermostatControl} introduces:

\begin{itemize}
  \item new categories (\texttt{Location}, \texttt{Temperature},
        \texttt{Action}),
  \item named constants (\texttt{office}, \texttt{celsius}),
  \item domain-specific predicates (\texttt{gtTemp}, \texttt{ltTemp}),
  \item and external actions (\texttt{lowerThermostat},
        \texttt{raiseThermostat}, \texttt{maintainThermostat}).
\end{itemize}

The same core could be reused for a completely different domain (for
example, cybersecurity incident response) by introducing new contexts
with categories such as \texttt{System}, \texttt{Incident},
\texttt{Recommendation}, and so on, but with the same underlying
boolean and tool machinery. This modularity reflects the separation
between universal grammar and domain-specific lexical items in
linguistic theory \citep{Chomsky1995}.

\subsection{Tools as Bridges Between Language and World}

The \texttt{Tool} structure in \texttt{Core1} is intentionally
generic. It can describe capabilities exposed via external protocols
such as web APIs or MCP tools \citep{Anthropic2024MCP}, but Prism 
itself does not prescribe how the underlying communication works.

From a linguistic viewpoint, \texttt{Tool} can be understood as a
bridge between the agent's internal policy language and the external
world. Role categories such as \texttt{Action} or \texttt{Response}
refine this general notion; they behave like semantic roles or verb
classes. A function that expects an \texttt{Action} is implicitly a
function that anticipates something executable, but the fine-grained
distinction between different roles prevents arbitrary re-use.

\section{Why a Language Like Prism is Useful}

There are several reasons to consider a small metalanguage like Prism
for agents:

\begin{itemize}
  \item \textbf{Inspectability.} Because agent policies are expressed
        as compact, compositional expressions, they can be inspected
        and reasoned about without delving into model internals.
  \item \textbf{Control and safety.} The set of allowed actions and
        tools is explicitly enumerated in contexts. This makes it
        easier to state, and check, safety constraints such as ``the
        agent may not call these tools under these conditions''.
  \item \textbf{Separation of concerns.} Linguistically rich prompts
        and context can still be handled by large language models, but
        the final decision over \emph{what to do} is constrained by a
        Prism policy. This suggests a hybrid architecture where
        natural language understanding and formal control coexist.
  \item \textbf{Reusability.} The same \texttt{Core1} can underlie
        very different domains; contexts serve as small, reusable
        grammars of agent behaviour for specific applications.
  \item \textbf{Verifiability.} The formal structure allows for
        automated verification of properties such as ``no critical
        action is taken without proper authorization'' or ``all sensor
        readings are validated before use''.
\end{itemize}

For linguists, Prism can be read as a proposal for a very small
interface language between semantic/pragmatic representations and
computational agents. It makes explicit the structure of decisions
and the vocabulary of tools, while staying agnostic about the details
of underlying algorithms.

\section{Conclusion}

Prism is a minimalist metalanguage for agent behaviour. It is built
out of a tiny core of abstract combinators (booleans, predicates,
pairs, tools) and a notion of contexts that introduce domain-specific
categories and external capabilities. Decisions that might usually be
expressed with imperative control flow are here written as ordinary
expressions that select among alternatives.

This design is motivated by both linguistic and engineering concerns:
we want agent behaviour to be compositional, lexically transparent,
and formally inspectable, while still rich enough to describe
real-world tasks. The examples in this paper demonstrate how Prism
can specify diverse agent behaviours---from thermostat control to
medical monitoring---using the same underlying compositional machinery.

Future work could connect Prism more directly to existing frameworks
in formal semantics and dialogue modelling, and explore how linguists
might use Prism-like languages to specify the behaviour of
tool-using agents in a principled way. Additionally, developing
tooling for automated verification and policy synthesis would make
Prism more practical for deployment in safety-critical applications.

\bibliographystyle{plainnat}

\end{document}